\documentclass[letterpaper,10pt,conference]{ieeeconf}

\usepackage{graphicx}
\usepackage{amsmath}
\usepackage{amsfonts}
\usepackage{booktabs}
\usepackage[english]{babel}

\IEEEoverridecommandlockouts
\newif\ifRALanonymous
\RALanonymousfalse

\title{\LARGE \bf Brace Yourself: Task-Conditioned Environmental Bracing \\for Forceful Humanoid Manipulation}

\ifRALanonymous
\author{\mbox{}}
\else
\author{Zongyuan Zhang$^{1,2,3,*}$, Christopher Lehnert$^{1,2,3}$, Will N.
Browne$^{1,2,3}$, and Jonathan M. Roberts$^{1,2,3}$%
\thanks{$^{1}$School of Electrical Engineering and Robotics, Queensland
University of Technology, 2 George St, Brisbane, 4000, Queensland, Australia.}%
\thanks{$^{2}$Australian Cobotics Centre, Queensland University of Technology,
2 George St, Brisbane, 4000, Queensland, Australia.}%
\thanks{$^{3}$Centre for Robotics, Queensland University of Technology,
2 George St, Brisbane, 4000, Queensland, Australia.}%
\thanks{$^{*}$Corresponding author: {\tt\small z203.zhang@hdr.qut.edu.au}.}%
}
\fi

\begin{document}
\bstctlcite{bstctl:forced_etal,bstctl:nodash}
\maketitle
\thispagestyle{empty}
\pagestyle{empty}

\begin{abstract}

Forceful manipulation is challenging for humanoid robots because interaction forces can disturb whole-body balance. We introduce the Supporting Hand Strategy (SHS), which enables a humanoid to brace against the environment with one hand while performing forceful manipulation with the other. SHS optimises a task-conditioned support configuration that guides two synchronous reinforcement-learning policies, without human motion data or online whole-body trajectory planning. On a Unitree G1, SHS achieved usable contact forces up to 60\,N, compared with a maximum sustained force of 13.5\,N without environmental bracing, while substantially improving force tracking over a task-independent support configuration. The same policies generalised to different task regions without retraining. SHS therefore provides a simple mechanism for substantially extending humanoid forceful-manipulation capability.

\end{abstract}

\section{Introduction}

Forceful manipulation poses a fundamental challenge for humanoid robots. Unlike fixed-base manipulators, a humanoid is not rigidly attached to the environment: interaction forces applied by the hand can propagate through the kinematic chain, disturb whole-body balance, and displace the floating base. As the required task force increases, this coupling can reduce position accuracy and ultimately limit the force that the robot can sustain. These effects are particularly important in contact-rich tasks such as cutting, grinding, polishing, pushing, and assembly, where successful manipulation requires simultaneous control of both motion and interaction force.

Humans commonly address the same physical problem by using the environment for additional support. During forceful bimanual activity, one hand may perform the task while the other braces against a nearby surface to stabilise the body~\cite{Woytowicz2018-bm}. This additional contact changes the mechanical support available to the body and can help resist reaction forces generated at the working hand. We investigate whether the same principle can be exploited by a humanoid robot to extend its forceful manipulation capability.

\begin{figure}[t!]
  \centering
  \includegraphics[width=\columnwidth]{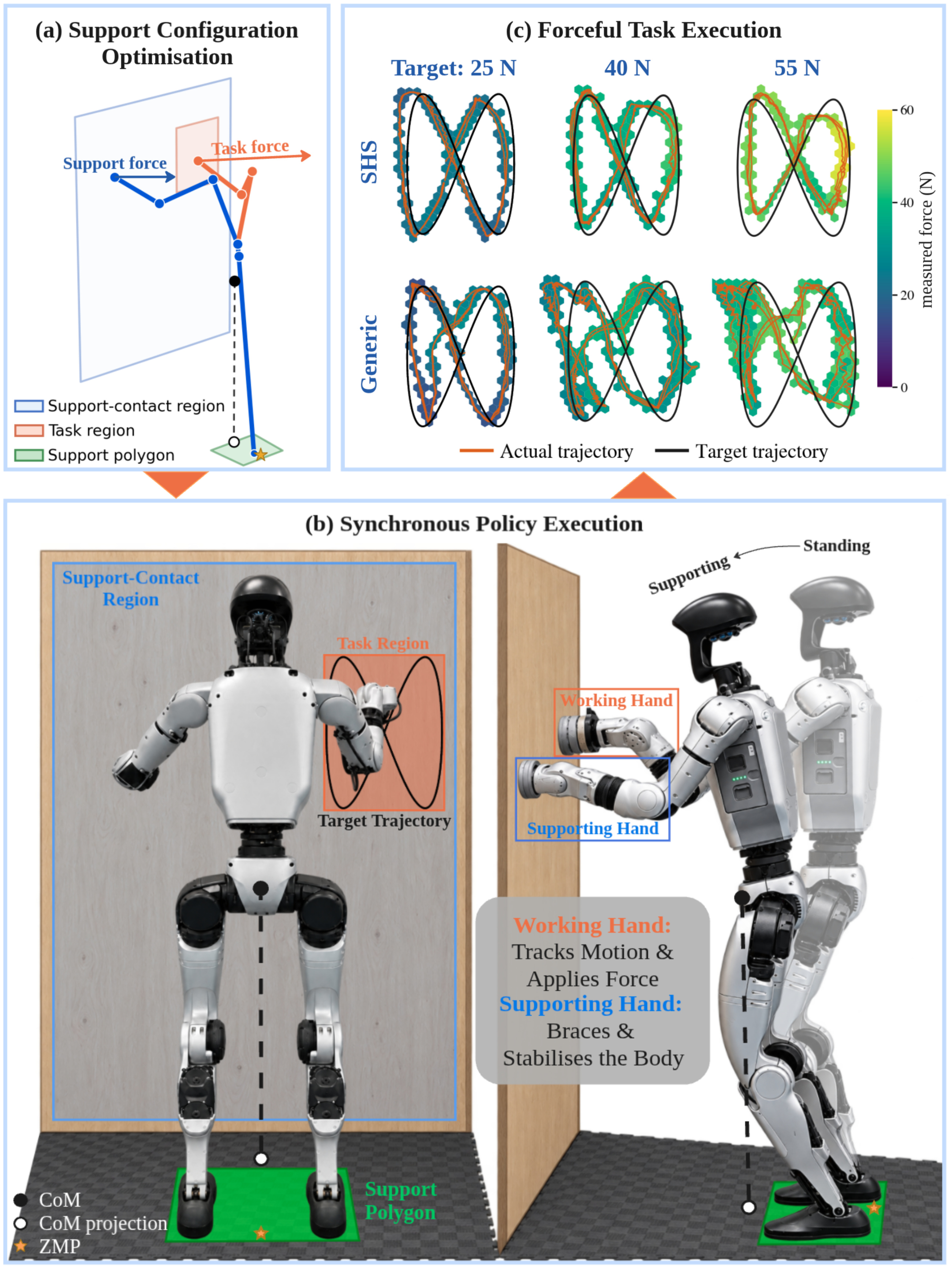}
  \caption{Supporting Hand Strategy (SHS). (a) Support configuration optimisation. Given the working-hand task region and the admissible support-contact region, a support configuration is optimised so that the task force applied by the working hand and the support force at the supporting hand keep the zero-moment point (ZMP) within the support polygon. (b) During execution, the body policy establishes and maintains environmental bracing, while the working-arm policy tracks a target position--force trajectory; the legend marks the whole-body centre of mass (CoM), its ground projection, and the ZMP. (c) Hardware figure-eight tracking at target forces of 25, 40, and 55\,N, with marker colour showing the measured force. Task-conditioned SHS follows the target trajectory more closely than the Generic variant that uses a task-independent reference.}
  \label{fig:overview}
\end{figure}

We propose the \emph{Supporting Hand Strategy} (SHS), which enables a humanoid to brace against the environment with one hand while performing forceful manipulation with the other. Rather than imitating a demonstrated whole-body motion~\cite{He2024-oh}, SHS captures the functional principle of bracing and determines how the robot should brace for a given manipulation task. As illustrated in Fig.~\ref{fig:overview}, SHS first optimises a support configuration from the working-hand position--force requirements across a task region \emph{and} an admissible support-contact region. Crucially, this optimisation produces a \emph{single} task-conditioned support configuration that is suitable across the complete task region, rather than a full-body trajectory that must be planned online.

The optimised support configuration is encoded as a reference for two synchronous reinforcement-learning (RL) policies. A body policy controls the legs, waist, and supporting arm to establish and maintain environmental bracing, while a separate working-arm policy tracks the commanded position--force trajectory. This separation allows the support configuration to remain stable while the working arm moves independently through the task region. SHS therefore requires neither human motion data nor predefined or online-generated whole-body trajectories, and neither policy requires measured force feedback during execution.

Our experiments isolate three questions: whether environmental bracing itself is necessary, whether the learned controller requires an explicit support reference, and whether that reference should be conditioned on the manipulation task. We evaluate these questions through simulation ablations and hardware experiments on a Unitree G1. We additionally compare SHS with a conventional unsupported baseline that combines the robot's lower-body RL controller with inverse kinematics (IK) for the working arm. The resulting experiments show that environmental bracing substantially widens the usable force range, while task-conditioned support configuration further improves position--force tracking. The same RL policies also execute newly optimised support configurations in shifted task regions without retraining.

The contributions of this work are:

\noindent\textbf{1) Supporting Hand Strategy:} to the best of our knowledge, this is the first strategy for floating-base humanoids to use one hand for environmental bracing specifically to increase the forceful manipulation capability of the other hand;

\noindent\textbf{2) Task-conditioned support optimisation:} an offline method for finding a single support configuration suitable across a manipulation task region and using it to guide synchronous body and working-arm policies without human motion data or online whole-body trajectory planning;

\noindent\textbf{3) Simulation and real-robot validation:} experiments that isolate the effects of environmental bracing, support-reference guidance, and task conditioning, and demonstrate increased usable force, improved high-force tracking, and transfer to shifted task regions without policy retraining.

\section{Related Work}

Contact-rich manipulation requires simultaneous regulation of motion and interaction force~\cite{Suomalainen2022-ih}. For fixed-base manipulators, hybrid position--force and impedance control provide the standard foundations for coordinating these objectives~\cite{Raibert1981-il,Hogan1985-rk}, and have been widely applied to forceful tasks such as robotic grinding and sanding~\cite{robotic_sanding_overview,Maric2020-qm}. Humanoid robots introduce an additional challenge because their floating base couples interaction forces with whole-body balance. Zero-moment point (ZMP)-based methods provide a foundation for humanoid balance~\cite{Vukobratovic2004-zm,Kajita2003-bp}, while operational-space control and hierarchical whole-body optimisation coordinate manipulation objectives with whole-body dynamics and contact constraints~\cite{Khatib1987-os,Escande2014-hq}. Partial force control similarly regulates whole-body motion together with selected contact forces on floating-base robots~\cite{DelPrete2014-pfc}, and load-dependent impedance regulation and operator feedback have been used to manage destabilising interaction forces during humanoid telemanipulation~\cite{Brygo2014-teleop}. These approaches regulate motion, force, and balance using the available contacts. When support is provided only by the feet, however, the sustainable interaction force remains constrained by the foot support polygon and the robot's ability to resist the reaction force. Establishing additional contact with the environment provides a direct way to alter these constraints.

Environmental bracing has a long history in robotic manipulation. Book et al. formulated a bracing strategy for robot operation~\cite{Book1984-bracing}, Fang et al. exploited non-end-effector environmental contacts to reduce manipulator joint effort~\cite{Fang2019-support}, and Wang and Minami used multiple intermediate-link contacts on a mobile manipulator to improve accuracy and reduce energy consumption~\cite{Wang2010-bracing}. These works demonstrate the mechanical benefits of exploiting environmental contacts, but concern fixed- or mobile-base manipulators rather than bipedal floating-base humanoids. In humanoid robotics, multi-contact methods have addressed passivity-based balancing~\cite{Henze2016-bw}, whole-body contact-force control~\cite{Rouxel2024-se}, and planning and control for multi-contact manipulation~\cite{Polverini2020-mx}. Together, these works establish methods for controlling and planning behaviour once multiple contacts are available. They do not, however, address the task-level problem considered here: determining a single support configuration that remains suitable across the position--force requirements of an entire manipulation task region. Moreover, these approaches are predominantly model-based and rely on whole-body dynamics and contact information, increasing modelling and sensing requirements during hardware deployment.

\begin{figure*}[t]
  \centering
  \begingroup
  \setlength{\unitlength}{\textwidth}
  \begin{picture}(1,0.294720)
    \put(0,0){\includegraphics[width=\textwidth]{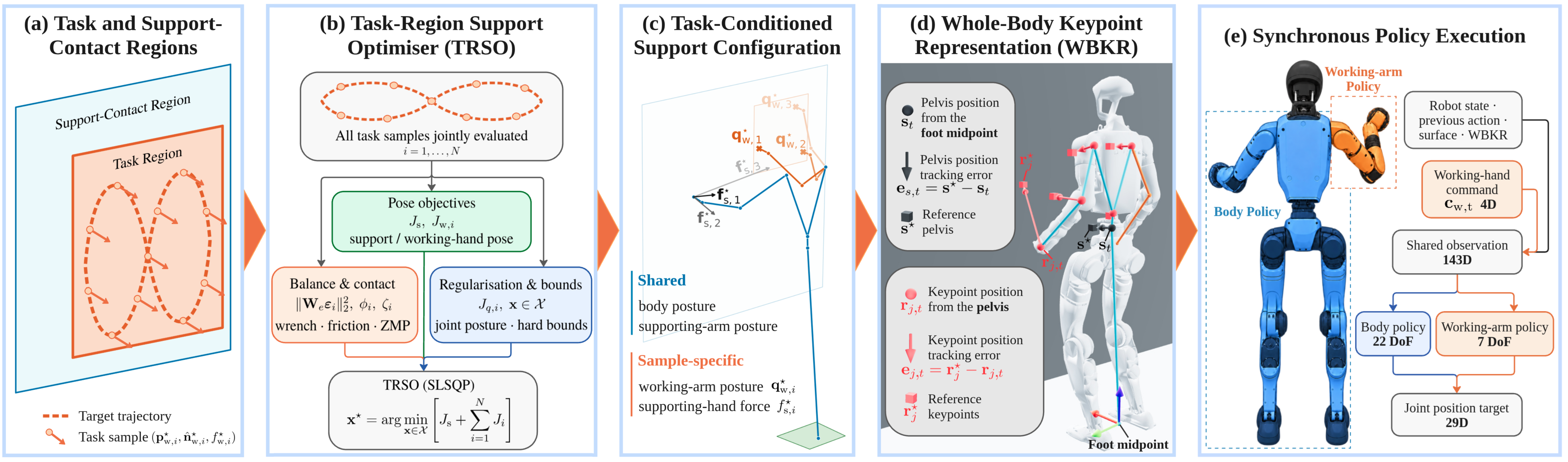}}
    \put(0.730,0.030){\makebox(0,0){\tiny $\{H\}$}}
  \end{picture}
  \endgroup
  \caption{SHS pipeline. (a) The task region is discretised into task samples for the working hand, while the support-contact region defines admissible contacts for the supporting hand. (b) The Task-Region Support Optimiser jointly optimises the support configuration across all task samples. (c) The support configuration specifies the shared body and supporting-arm posture, while the working-arm posture and supporting-hand contact force vary across task samples. (d) The Whole-Body Keypoint Representation (WBKR) encodes reference--measurement errors using the pelvis position relative to the foot midpoint and body and supporting-arm keypoint positions relative to the pelvis, in frame $\{H\}$. (e) Synchronous working-arm and body policies share the same observation and generate whole-body joint-position targets to track the target position--force trajectory while maintaining environmental bracing.}
  \label{fig:method}
\end{figure*}

Learning-based humanoid control offers an alternative to explicitly modelling all aspects of whole-body interaction. Learned controllers have demonstrated sequential-contact and multi-support behaviours~\cite{Zhang2024-wo,Rouxel2024-fm}, tactile whole-body contact manipulation~\cite{Murooka2025-tc}, compliant control~\cite{Margolis2025-by}, and force-adaptive loco-manipulation~\cite{Zhang2025-fa}. These results show that learned humanoid controllers can accommodate diverse contact and force interactions, but environmental bracing has not been formulated as a task-level mechanism for extending sustainable manipulation force. Furthermore, learned whole-body controllers commonly require a reference configuration or motion to realise the desired behaviour. Demonstration- and teleoperation-based approaches obtain such references from human motion and embodiment retargeting~\cite{He2024-oh}, while model-based pipelines can generate dynamically feasible whole-body trajectories for a learned policy to track~\cite{Liu2025-o2s}. The former depends on the quality and coverage of human demonstrations and must accommodate human--robot morphology differences, while the latter requires generation of complete dynamically feasible whole-body trajectories.

This motivates a different formulation: rather than reproducing a demonstrated motion or generating a complete whole-body trajectory, the robot can determine how to brace for the manipulation task and use the resulting support configuration to guide learned execution. What remains missing is a demonstration-free method for determining a task-conditioned support configuration that remains suitable across the position--force requirements of an entire manipulation task, while allowing a learned controller to execute the task without online whole-body trajectory generation or measured force feedback, thus simplifying deployment and potentially improving forceful-manipulation performance.

\section{Supporting Hand Strategy}
\label{sec:method}

SHS separates forceful manipulation into offline support-configuration optimisation and RL policy execution. It uses three components: the Task-Region Support Optimiser (TRSO), the Whole-Body Keypoint Representation (WBKR), and synchronous body and working-arm policies. TRSO finds one support configuration that accommodates the working-hand position--force requirements across a task region while placing the supporting hand within an admissible support-contact region. WBKR encodes this configuration geometrically for policy input. During execution, the body policy maintains bracing while the working-arm policy tracks the position--force command (Fig.~\ref{fig:method}).

\subsection{Task and Support-Contact Regions}

The task region is an area on the task surface. It is discretised into a set $\mathcal T$ of $N$ task samples specifying the working-hand requirements (Fig.~\ref{fig:method}(a)):
\begin{equation}
\mathcal{T}=\{(\mathbf{p}^{\star}_{\mathrm{w},i},
\hat{\mathbf{n}}^{\star}_{\mathrm{w},i},
f^{\star}_{\mathrm{w},i})\}_{i=1}^{N}.
\label{eq:regions}
\end{equation}
Here $\mathbf p^{\star}_{\mathrm{w},i}$, $\hat{\mathbf n}^{\star}_{\mathrm{w},i}$, and $f^{\star}_{\mathrm{w},i}$ denote target position, task-surface normal, and target normal-force magnitude for the working hand, respectively. The support-contact region $\mathcal S$ is a continuous rectangle that specifies where bracing is permitted on the planar support surface with unit normal $\hat{\mathbf n}_{\mathcal S}$. Throughout, bold symbols denote vectors and plain symbols scalar magnitudes, and a star marks a target, reference, or optimised value as opposed to a measured one.

Crucially, SHS optimises a \emph{single} support configuration, and hence a single supporting-hand contact point, for the entire task region rather than planning a sequence of whole-body configurations along the working-hand trajectory. The planar position of the foot midpoint, the forward lean that pitches the body towards the task surface, waist posture, and supporting-arm posture are therefore shared across all task samples. The working-arm posture and supporting-hand contact force may vary with sample $i$ so that each position--force requirement can be evaluated under the same candidate support configuration. This trades per-sample support optimisation for avoiding online replanning throughout the task.

\subsection{Task-Region Support Optimiser}

TRSO searches for a common support configuration that accommodates the position--force requirements across the task region while maintaining kinematic feasibility and quasi-static balance; it does not plan a whole-body trajectory.

Given $\mathcal{T}$ and $\mathcal{S}$, TRSO solves an offline constrained nonlinear optimisation problem (Fig.~\ref{fig:method}(b,c)). Its decision variables comprise a 14D support configuration $\mathbf x_{\mathrm{s}}$ shared across all task samples and $N$ sample-specific 8D variables $\mathbf x_i$, allowing the working-arm configuration and supporting-hand force to vary across task samples. $\mathbf x_{\mathrm{s}}$ contains the planar position $(b_x,b_y)$ of the foot midpoint, ankle pitch $q_{\mathrm{ankle}}$, pelvis pitch $q_{\mathrm{pelvis}}$, waist joints $\mathbf{q}_{\mathrm{waist}}\in\mathbb{R}^3$, and supporting-arm joints $\mathbf q_{\mathrm{s}}\in\mathbb R^7$. Each $\mathbf x_i$ contains the working-arm joints $\mathbf q_{\mathrm{w},i}\in\mathbb R^7$ and the supporting-hand normal force $f_{\mathrm{s},n,i}$:
\begin{equation}
\begin{aligned}
\mathbf x_{\mathrm{s}}&=[b_x,b_y,q_{\mathrm{ankle}},q_{\mathrm{pelvis}},
\mathbf q_{\mathrm{waist}}^{\top},
\mathbf q_{\mathrm{s}}^{\top}]^{\top}\in\mathbb R^{14},\\
\mathbf x_i&=[\mathbf q_{\mathrm{w},i}^{\top},
f_{\mathrm{s},n,i}]^{\top}\in\mathbb R^8.
\end{aligned}
\label{eq:decision}
\end{equation}
Stacking gives the decision vector $\mathbf x=[\mathbf x_{\mathrm{s}}^{\top},\mathbf x_1^{\top},\ldots,\mathbf x_N^{\top}]^{\top}\in\mathbb R^{14+8N}$.

For each task sample, forward kinematics (FK) gives the hand poses and whole-body centre of mass. Quasi-static evaluation uses the whole-body gravitational load $\mathbf f_{\mathrm{g}}$, the prescribed working-hand force $\mathbf f_{\mathrm{w},i}$, which is modelled as purely normal to the task surface, and the supporting-hand force $\mathbf f_{\mathrm{s},i}$, which may also transmit a tangential component:
\begin{equation}
\begin{aligned}
\mathbf f_{\mathrm{w},i}&=f^{\star}_{\mathrm{w},i}
\hat{\mathbf n}^{\star}_{\mathrm{w},i},\\
\mathbf f_{\mathrm{s},i}&=f_{\mathrm{s},n,i}
\hat{\mathbf n}_{\mathcal S}+\mathbf f_{\mathrm{s},t,i},\\
\boldsymbol{\varepsilon}_{i}
&=[\mathbf F_i^{\top},\boldsymbol\tau_i^{\top}]^{\top}
\in\mathbb R^6,
\end{aligned}
\label{eq:quasistatic}
\end{equation}
where $\mathbf F_i=\mathbf f_{\mathrm{g}}+\mathbf f_{\mathrm{w},i}+\mathbf f_{\mathrm{s},i}$ and $\boldsymbol\tau_i$ are the resultant force and moment about the ground projection of the foot midpoint $\mathbf p_0=[b_x,b_y,0]^\top$, so $\boldsymbol\varepsilon_i$ is the residual wrench remaining for the ground contact to balance. The tangential force $\mathbf f_{\mathrm{s},t,i}$ follows from least-squares moment balance, rescaled to the Coulomb friction bound. Force balance gives the ground reaction as $-\mathbf F_i$, which is used together with $\boldsymbol\tau_i$ to compute the ZMP.

TRSO also evaluates the supporting-hand pose against the support-contact region, using a position term $e_{p,\mathrm{s}}$ that keeps the hand on the surface and within the admissible region $\mathcal S$, and an orientation term $e_{o,\mathrm{s}}$ that penalises misalignment between the palm normal and $\hat{\mathbf n}_{\mathcal S}$. With wrench-weighting matrix $\mathbf W_e$, the objective is
\begin{equation}
\begin{aligned}
\mathbf{x}^{\star}
&=\arg\min_{\mathbf{x}\in\mathcal X}
\left[J_{\mathrm{s}}+\sum_{i=1}^{N}J_i\right],\\
J_i
&=J_{\mathrm{w},i}
+\lambda_e\|\mathbf W_e\boldsymbol\varepsilon_i\|_2^2
+\lambda_\phi\phi_i+\lambda_z\zeta_i+\lambda_qJ_{q,i}.
\end{aligned}
\label{eq:objective}
\end{equation}
$J_{\mathrm{s}}$ combines the supporting-hand errors $e_{p,\mathrm{s}}$ and $e_{o,\mathrm{s}}$ with regularisation of the shared foot-midpoint position and body posture, while $J_{\mathrm{w},i}$ combines the working-hand position and palm-orientation errors. The friction term $\phi_i$ penalises ground-friction excess implied by the required reaction $-\mathbf F_i$ and supporting-hand friction utilisation, while $\zeta_i$ penalises ZMP excursion outside the foot support polygon, approximated as a rectangle. $J_{q,i}$ regularises joint displacement, and the $\lambda$ terms are scalar cost weights. Joint limits and bounds on the planar position of the foot midpoint define the box-constraint set $\mathcal X$. Apart from the friction limit on $\mathbf f_{\mathrm{s},t,i}$, all terms are smooth penalties.

Because the floor supplies the ground normal reaction, the vertical component of $\boldsymbol\varepsilon_i$ has zero weight in the wrench-weighting matrix $\mathbf W_e$, while the horizontal components regularise the required horizontal ground reaction force. The supporting-hand normal force is bounded to non-negative values, and the program is solved with a sequential least-squares quadratic programming (SLSQP) solver, which returns a local solution rather than a global optimum.

\subsection{Whole-Body Keypoint Representation}

The WBKR maps the support configuration optimised by TRSO to geometric references for policy input (Fig.~\ref{fig:method}(d)). FK then extracts eight three-dimensional keypoints from this configuration: the pelvis, torso, both shoulders, both elbows, and both hands, yielding a 24D representation.

WBKR represents the configuration using relative geometric offsets about the pelvis. Let $\mathbf p_{\mathrm{foot}}$ denote the midpoint of the two feet. All vectors use the task-aligned heading frame $\{H\}$ marked in Fig.~\ref{fig:method}(d), whose orientation is fixed at task activation: its horizontal axes are aligned with the initial pelvis yaw, with zero roll and pitch. This makes the representation invariant to the robot's initial global yaw while keeping the WBKR reference coordinates constant throughout execution. WBKR is defined by
\begin{equation}
\begin{aligned}
\mathbf{s}&=\mathbf{p}^{H}_{\mathrm{pel}}-
\mathbf{p}^{H}_{\mathrm{foot}},\\
\mathbf{r}_{j}&=\mathbf{p}^{H}_{j}-
\mathbf{p}^{H}_{\mathrm{pel}}.
\end{aligned}
\label{eq:representation}
\end{equation}
Here, $\mathbf{s}$ is the pelvis support vector from the foot midpoint to the pelvis, while $\mathbf{r}_j$ describes the position of body keypoint $j$ relative to the pelvis. The same representation is applied to the task-conditioned WBKR reference $(\mathbf{s}^{\star},\mathbf{r}^{\star}_{j})$ and the current robot state $(\mathbf{s}_{t},\mathbf{r}_{j,t})$. This decomposition keeps the leg joints out of $\mathbf{r}_j$: the stance geometry that produces the forward lean is carried by $\mathbf{s}$ alone, so that the upper-body posture target does not depend on leg tracking error.

The policy reference $\mathbf k^{\star}$ retains six support-configuration keypoints: the pelvis, torso, both shoulders, supporting elbow, and supporting hand,
\begin{equation}
\mathbf{k}^{\star}=\left[
\mathbf{s}^{\star\top},
\mathbf{r}^{\star\top}_{\mathrm{tor}},
\mathbf{r}^{\star\top}_{\mathrm{s,sh}},
\mathbf{r}^{\star\top}_{\mathrm{w,sh}},
\mathbf{r}^{\star\top}_{\mathrm{s,el}},
\mathbf{r}^{\star\top}_{\mathrm{s,hand}}
\right]^{\top}\in\mathbb{R}^{18}.
\label{eq:keypoint-reference}
\end{equation}
The working elbow and hand are excluded so that the working arm remains free to track its separate position--force command. The measured WBKR $\mathbf{k}_{t}\in\mathbb{R}^{21}$ is constructed analogously from the current robot state with the current working-hand position additionally included.

\subsection{Synchronous Dual-Policy Learning and Execution}
\label{sec:dualpolicy}

The two policies separate bracing maintenance from working-hand trajectory tracking. The working-arm policy $\pi_{\mathrm{w}}$ controls all seven working-arm joints, while the body policy $\pi_{\mathrm{b}}$ controls both legs, the waist, and the supporting arm (Fig.~\ref{fig:method}(e)). Both map a shared observation $\mathbf{o}_t$ to actions:
\begin{equation}
\begin{aligned}
\mathbf{a}_{\mathrm{w},t}&=\pi_{\mathrm{w}}(\mathbf{o}_{t})\in\mathbb{R}^{7},\\
\mathbf{a}_{\mathrm{b},t}&=\pi_{\mathrm{b}}(\mathbf{o}_{t})\in\mathbb{R}^{22}.
\end{aligned}
\label{eq:policies}
\end{equation}
The combined action is $\mathbf{a}_{t}=[\mathbf{a}_{\mathrm{w},t}^{\top},\mathbf{a}_{\mathrm{b},t}^{\top}]^{\top}\in\mathbb{R}^{29}$. The two policies control disjoint joint sets and operate synchronously rather than hierarchically. 

The proprioceptive state and working-hand command are
\begin{equation}
\begin{aligned}
\mathbf{u}_{t}&=\left[\boldsymbol{\omega}_{\mathrm{b},t}^{\top},\mathbf{g}_{\mathrm{b},t}^{\top},(\mathbf{q}_{t}-\mathbf{q}_{0})^{\top},\dot{\mathbf{q}}_{t}^{\top}\right]^{\top}\in\mathbb{R}^{64},\\
\mathbf{c}_{\mathrm{w},t}&=\left[(\mathbf{p}^{H,\star}_{\mathrm{w},t}-\mathbf{p}^{H,\star}_{\mathrm{pel}})^{\top},f^{\star}_{\mathrm{w},t}\right]^{\top}\in\mathbb{R}^{4}.
\end{aligned}
\label{eq:policy-inputs}
\end{equation}
Here $\boldsymbol{\omega}_{\mathrm{b},t}$ is the base angular velocity, $\mathbf{g}_{\mathrm{b},t}$ projected gravity, $\mathbf q_t$ and $\dot{\mathbf q}_t$ the joint positions and velocities, $\mathbf q_0$ the default posture, $\mathbf{p}^{H,\star}_{\mathrm{w},t}$ the target working-hand position, and $\mathbf{p}^{H,\star}_{\mathrm{pel}}$ the task-conditioned pelvis reference.

Let $\hat d$ denote the surface-distance estimate calibrated at task activation, $\mathbf v_{\mathrm{w,sh},t}$ the working-shoulder velocity relative to the pelvis, and $\mathbf a_{t-1}\in\mathbb R^{29}$ the combined previous action. The shared actor observation $\mathbf o_t\in\mathbb R^{143}$ is
\begin{equation}
\mathbf{o}_{t}=\left[
\mathbf{u}_{t}^{\top},
\mathbf{a}_{t-1}^{\top},
\mathbf{c}_{\mathrm{w},t}^{\top},
\hat{\mathbf n}_{\mathcal S}^{\top},
\hat d,
\mathbf{k}^{\star\top},
\mathbf{k}_{t}^{\top},
\mathbf{v}_{\mathrm{w,sh},t}^{\top}
\right]^{\top}.
\label{eq:actor-observation}
\end{equation}
$\mathbf{o}_t$ includes the combined previous action, facilitating coordination between working-hand motion and environmental bracing. The actors receive the target working-hand force but no measured force or contact quantities, relying on proprioception rather than force feedback to regulate the contact force; these privileged measurements are used only by the critics and reward functions during training. The two actors have separate parameters and observation normalisers.

The combined action is mapped to joint-position targets as a per-joint scaled offset from the default posture $\mathbf q_0$, tracked by joint proportional--derivative control.

The reward functions reflect the distinct policy roles:
\begin{equation}
\begin{aligned}
r_{\mathrm{w},t}&=\lambda_p r_{\mathrm{w,pose}}
+\lambda_f r_{\mathrm{w,force}}
-\lambda_{\tau,\mathrm{w}}\rho_{\mathrm{w}}
-\mathcal R_{\mathrm{w}},\\
r_{\mathrm{b},t}&=\lambda_k r_{\mathrm{WBKR}}
+\lambda_c r_{\mathrm{s,contact}}
-\lambda_{f,\mathrm{s}}\rho_{\mathrm{s,force}}\\
&\quad-\lambda_b\rho_{\mathrm{foot}}
-\lambda_{\tau,\mathrm{s}}\rho_{\mathrm{s}}
-\mathcal R_{\mathrm{b}}.
\end{aligned}
\label{eq:policy-rewards}
\end{equation}
The $\lambda$ terms are scalar reward weights. The working-arm reward tracks the commanded hand pose and normal force through $r_{\mathrm{w,pose}}$ and $r_{\mathrm{w,force}}$. The body-policy reward uses $r_{\mathrm{WBKR}}$ to track the pelvis support vector and pelvis-relative upper-body geometry. Since no supporting-hand force is prescribed, $r_{\mathrm{s,contact}}$ rewards supporting-hand contact while $\rho_{\mathrm{s,force}}$ discourages unnecessary supporting-hand force. The term $\rho_{\mathrm{foot}}$ regularises inter-foot load sharing, within-foot pressure distribution, and their temporal variation. $\mathcal R_{\mathrm{w}}$ and $\mathcal R_{\mathrm{b}}$ collect the remaining weighted regularisation terms, including action smoothness, joint-limit violation, self-collision, and hand-contact impact.

Early hardware tests showed that the shoulder-pitch actuator was consistently the first joint on either arm to approach its thermal limit. Prior work addresses thermal limits directly by adapting joint torque limits from online temperature estimation~\cite{Kumagai2014TemperatureTorqueLimit}; instead, we penalise torque utilisation:
\begin{equation}
\rho_r=
\left[
\frac{1}{|\mathcal A_r|}
\sum_{j\in\mathcal A_r}
\left(\frac{|\tau_j|}{\tau_j^{\max}}\right)^4
\right]^{1/2},
\qquad r\in\{\mathrm{w},\mathrm{s}\},
\label{eq:torque-utilisation}
\end{equation}
where $\mathcal A_r$ is the joint set of arm $r$. The fourth power penalises highly utilised joints disproportionately, discouraging sustained load concentration on a single actuator.

The policies are learned with proximal policy optimisation (PPO) in two stages: both are first trained on static targets, after which the body policy is frozen while the working-arm policy is adapted to continuous position--force commands. 

\section{Experimental Design}

\subsection{Baselines and Ablations}

The ablations isolate three components of SHS: task-conditioned support configuration, WBKR guidance, and environmental bracing. Table~\ref{tab:methods} summarises the comparisons. The four RL methods share the same dual-policy architecture, task distribution, dynamics, PPO settings, and training budget; ablated inputs and reward terms differ only as described below.

SHS uses the task-conditioned WBKR reference $\mathbf{k}^{\star}$ obtained from TRSO, whereas the benchmark, termed Generic, uses a single training-set medoid WBKR reference across all task regions, defined as the reference closest to the mean in the normalised 18D WBKR space. No-reference removes $\mathbf{k}^{\star}$ and the WBKR-tracking reward $r_{\mathrm{WBKR}}$. No-bracing additionally removes the supporting-hand contact reward and penalises unintended support-surface contact. To control for reachability, it retains a task-matched forward lean while receiving no WBKR reference.

Modular provides a hardware baseline that combines the robot's onboard lower-body RL controller with IK for the working arm. It follows the same working-hand commands without supporting-hand contact, and a proportional--integral force loop adjusts the hand target along the task-surface normal from the measured normal-force error, which is then realised by IK.

\begin{table}[t]
\caption{Baseline and ablation methods used in the experiments.}
\label{tab:methods}
\centering
\scriptsize
\setlength{\tabcolsep}{3.0pt}
\begin{tabular}{@{}lccll@{}}
\toprule
Method & Bracing & WBKR ref. & Reference source & Experimental role \\
\midrule
SHS & yes & yes & Task-conditioned & Complete method \\
Generic & yes & yes & Training-set medoid & Task conditioning \\
No-reference & yes & no & -- & WBKR guidance \\
No-bracing & no & no & -- & Environmental bracing \\
Modular & no & no & -- & Hardware baseline \\
\bottomrule
\end{tabular}
\end{table}

\subsection{Data and Training}
\label{sec:training}

Task regions vary in surface distance, working-hand position and force, and support-contact-region geometry. Each region contains 30 task samples over 0--75\,N, jointly optimised by TRSO to produce one support configuration. The dataset contains 1,680 training and 336 held-out configurations (50,400 training samples), split by support configuration.

Training uses two stages. Stage one trains both policies on static targets for 9,000 PPO updates. Stage two freezes the body policy and its observation normaliser and trains only the working-arm policy for a further 3,000 updates using static-position variable-force scenarios and figure-eight trajectories at constant and variable force, with 35\% static-target replay. Figure-eight trajectories vary motion direction, velocity, and acceleration, testing tracking under changing dynamic loads.

During stage one, a curriculum initially disables the $\rho_r$ penalties, then ramps their weights once smoothed hand-tracking errors reach preset thresholds, so that the policies first learn to complete the task and then learn to lower the actuator load. For all RL methods, we select the checkpoint with the lowest normalised position--force training-error score after completion of the curriculum. Domain randomisation covers hand--surface friction, contact compliance, support-surface placement, surface-distance calibration residual, observation noise, and base-attitude drift, representing key sources of sim-to-real mismatch for this task.

\subsection{Evaluation Protocol and Metrics}

Simulation uses episode-level criteria. A stable-contact window begins after 0.1\,s of continuous contact and ends when contact breaks. Stable-contact hold ratio is the fraction of non-zero-force, on-surface steps within the stable-contact window, averaged over episodes that establish such a window. Success rate is averaged over all held-out episodes; success requires a stable-contact window, mean position error $\leq5$\,cm, normal-force mean absolute error (MAE) $\leq10$\,N within the window, and no early termination.

Two hardware experiments use a 29-DoF Unitree G1 facing a rigid vertical task surface. An Axia80-M20 six-axis force/torque sensor (ATI Industrial Automation) is mounted at the working wrist, while hand and shoulder positions are reconstructed by FK from joint encoders in the task-aligned heading frame $\{H\}$. For RL policies, measured force is used only for evaluation; Modular also uses it for feedback. Policies run at 50\,Hz and the joint-level controller at 200\,Hz.

\smallskip
\noindent\textbf{Experiment 1: Force capability at fixed positions}\par\nopagebreak
In the primary task region T1, the working hand is commanded to maintain a fixed target position (centre or one of four corners) while the target normal force is stepped through 0--80\,N in 10\,N levels, beyond the training force range. A level is sustained if the measured normal force remains between that level and the next, with standard deviation $\leq\max(2\,\mathrm{N},0.15\bar f_n)$, where $\bar f_n$ is the mean measured normal force. Each level is tested three times and is usable if sustained in at least two tests. Contiguous usable levels define the sustainable force range, the primary outcome.

\smallskip
\noindent\textbf{Experiment 2: Trajectory tracking}\par\nopagebreak
An eight-second figure-eight trajectory is scaled to cover as much of T1 as possible, with target forces of 10, 25, 40, 55, and 70\,N. Transfer tests translate T1 by 5\,cm laterally (T2) and vertically (T3) (Fig.~\ref{fig:transfer}(d)), using the same frozen SHS policies at 10, 25, 40, and 55\,N; 70\,N is excluded because T1 tracking already degrades.

For trajectory tracking, let $\hat{\mathbf n}$ be the task-surface normal, $\mathbf e_{p,t}$ the working-hand position error, $f_{n,t}$ the measured normal force, $f_{n,t}^{\star}$ the target normal force, and $\mathcal W$ the steady-state time steps. Normal-force and tangential-position root-mean-square (RMS) errors are
\begin{equation}
\begin{aligned}
e_f&=\sqrt{\frac{1}{|\mathcal W|}\sum_{t\in\mathcal W}
(f_{n,t}-f_{n,t}^{\star})^2},\\
e_{\mathrm{tan}}&=\sqrt{\frac{1}{|\mathcal W|}\sum_{t\in\mathcal W}
\left\|\mathbf e_{p,t}-
(\hat{\mathbf n}^{\top}\mathbf e_{p,t})\hat{\mathbf n}
\right\|_2^2}.
\end{aligned}
\label{eq:trajectory-errors}
\end{equation}
The primary outcome is T1 normal-force RMS error; tangential-position error is secondary. Exploratory working-shoulder repeatability is quantified by wobble, the RMS deviation from the cycle-aligned mean over $\geq3$ complete figure-eight cycles, and RMS linear speed in $\{H\}$ during steady-state contact. For a descriptive comparison of support configurations, $d_{\mathrm{med}}=\lVert\mathbf{k}^{\star}-\mathbf{k}^{\star}_{\mathrm{med}}\rVert_2$ measures the distance from the Generic reference in the 18D WBKR space.

Each trajectory recording is one data point at its target-force level. Simulation comparisons use 30 independently trained runs per method, paired by training seed, with paired Wilcoxon tests and paired-bootstrap 95\% confidence intervals. Hardware SHS--Generic comparisons use unadjusted two-sided Mann--Whitney tests separately at each force and Hedges' $g$, with positive $g$ indicating greater Generic error.

\section{Results}

\begin{table*}[!t]
\caption{Experiment 2: trajectory-tracking errors in primary task region T1; lower error is indicated in bold.}
\label{tab:t1-task-results}
\centering
\footnotesize
\setlength{\tabcolsep}{4.2pt}
\renewcommand{\arraystretch}{1.00}
\begin{tabular*}{\textwidth}{@{\extracolsep{\fill}}c*{10}{c}@{}}
\toprule
\multicolumn{1}{c}{\raisebox{-0.60\baselineskip}[0pt][0pt]{Target force (N)}} &
\multicolumn{4}{c}{Normal-force RMS error (N)} &
\multicolumn{4}{c}{Tangential-position RMS error (cm)} &
\multicolumn{2}{c}{Recordings} \\
\cmidrule(lr){2-5}\cmidrule(lr){6-9}\cmidrule(lr){10-11}
&
\multicolumn{1}{c}{Generic} & \multicolumn{1}{c}{SHS} &
\multicolumn{1}{c}{Hedges' $g$} & \multicolumn{1}{c}{$p$} &
\multicolumn{1}{c}{Generic} & \multicolumn{1}{c}{SHS} &
\multicolumn{1}{c}{Hedges' $g$} & \multicolumn{1}{c}{$p$} &
\multicolumn{1}{c}{Generic} & \multicolumn{1}{c}{SHS} \\
\midrule
10 & 4.6 & \textbf{3.3} & $2.17$ & $0.002$ & \textbf{3.2} & 3.3 & $-0.17$ & $0.818$ & 6 & 6 \\
25 & 7.9 & \textbf{3.3} & $1.90$ & $0.005$ & 3.3 & \textbf{3.1} & $0.26$ & $1.000$ & 6 & 7 \\
40 & 9.4 & \textbf{4.8} & $2.73$ & $0.004$ & 3.8 & \textbf{2.7} & $1.59$ & $0.052$ & 6 & 5 \\
55 & 13.1 & \textbf{9.2} & $2.99$ & $0.001$ & 5.1 & \textbf{3.6} & $1.74$ & $0.002$ & 6 & 7 \\
70 & 21.9 & \textbf{19.0} & $1.18$ & $0.048$ & 6.3 & \textbf{5.7} & $1.29$ & $0.019$ & 5 & 7 \\
\bottomrule
\end{tabular*}
\end{table*}

\subsection{Environmental Bracing Widens the Force Range}

\begin{figure}[t]
  \centering
  \includegraphics[width=\linewidth]{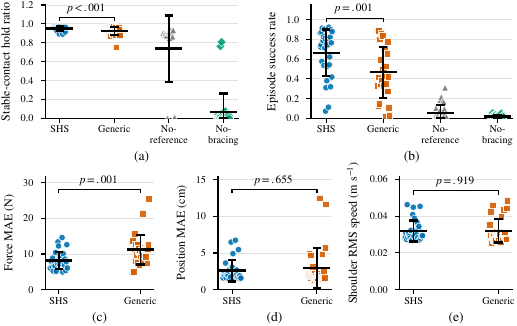}
  \caption{Simulation ablations over 30 independently trained runs. (a) Stable-contact hold ratio, (b) episode success rate, (c,d) contact-window force and position mean absolute errors (MAEs), and (e) working-shoulder root-mean-square (RMS) speed. Panel (a) includes runs that established a stable-contact window (No-reference, $n=17$; all others, $n=30$); (b) includes all runs; panels (c--e) compare SHS and Generic. Points denote runs, bars show mean $\pm$ standard deviation (SD), and brackets indicate paired Wilcoxon $p$-values. Higher is better in (a) and (b), lower is better in (c) and (d). Panel (e) is a control metric with no preferred direction.}
  \label{fig:simulation}
\end{figure}

Across 30 paired simulation runs, SHS outperformed No-bracing in every pair: environmental bracing raised episode success rate from $0.020$ to $0.666$ and mean stable-contact hold ratio from $0.069$ to $0.950$ (Fig.~\ref{fig:simulation}(a,b)).

In Experiment 1, SHS reached usable forces of 60\,N at the centre and upper positions and 40\,N at the lower positions, compared with 40--50\,N for Generic (Fig.~\ref{fig:hardware-results}(a)). Modular, the unsupported hardware baseline, was usable only at 10\,N at the lower-right position and never sustained more than 13.5\,N for one second.

The No-bracing simulation ablation failed differently: only two of 30 policies retained stable contact (hold ratios $0.763$ and $0.808$; all others $\leq0.074$), yet both produced $>21$\,N for an 8\,N target and retained load at zero target force. Without environmental bracing, the forward lean therefore routes body load through the working hand, imposing a non-zero minimum force. Environmental bracing avoids both this lower-force constraint and Modular's upper-force limitation.

\subsection{Task Conditioning Improves Tracking}

In simulation, SHS reduced force MAE by 3.02\,N (26.9\%), increased episode success rate by 0.199, and increased stable-contact hold ratio by 0.028 relative to Generic (paired Wilcoxon $p=0.001$, $p=0.001$, and $p<0.001$; Fig.~\ref{fig:simulation}(a--d)). Position MAE remained similar ($p=0.655$). The No-reference ablation rarely established reliable supporting-hand contact: 13 of 30 runs never established a stable-contact window, indicating that the contact reward alone was insufficient to guide the policy to establish environmental bracing under the tested training setting.

\begin{figure}[t]
  \centering
  \includegraphics[width=\linewidth]{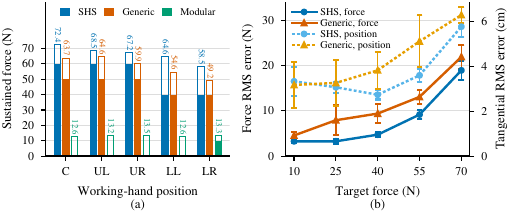}
  \caption{Hardware capability and tracking in T1. (a) Experiment 1: contiguous usable-force ranges (solid) and maximum one-second sustained forces (open) at the centre (C) and upper-left/right (UL/UR) and lower-left/right (LL/LR) corners. (b) Experiment 2: normal-force and tangential-position RMS errors across trajectory recordings at each force level (mean $\pm$ SD); lower is better.}
  \label{fig:hardware-results}
\end{figure}

In Experiment 2, SHS reduced normal-force RMS error at every tested force level. It approximately halved the normal-force RMS error at 25--40\,N and maintained lower error at higher loads (Fig.~\ref{fig:hardware-results}(b) and Table~\ref{tab:t1-task-results}). Differences in tangential-position error became more pronounced at higher loads, although both methods degraded at 70\,N. Neither policy observes measured force or force error, yet both track the commanded force without explicit force feedback.

\begin{figure}[t]
  \centering
  \includegraphics[width=\linewidth]{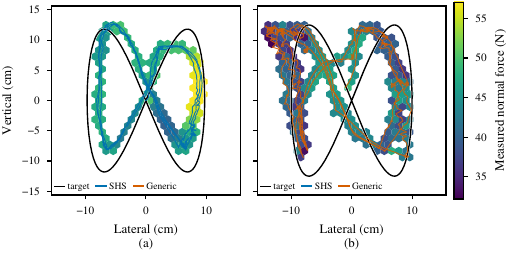}
  \caption{Matched hardware trajectory tracking in T1 at 55\,N: (a) SHS and (b) Generic. Black curves show the target trajectory; blue/orange curves show executed laps, with marker colour denoting mean measured normal force. At this load, SHS follows the target trajectory more closely than Generic.}
  \label{fig:trajectory}
\end{figure}

As shown in Fig.~\ref{fig:trajectory}, SHS achieved better trajectory tracking than Generic at 55\,N. Force RMS error decreased from 13.1 to 9.2\,N (30\%), and tangential-position RMS error from 5.1 to 3.6\,cm (29\%). In contrast, Modular accurately tracked only 11 of 29 recorded trajectories, all with target forces $\leq 11$\,N. For every recording above 11\,N, it either lost contact or maintained contact only partially.

\subsection{Task Conditioning Improves Shoulder Repeatability}

Lower working-shoulder wobble indicates more repeatable motion of the base of the working arm across trajectory cycles. In hardware trajectory tracking, SHS and Generic were similar at low loads, but SHS reduced wobble by factors of 2.7--3.8 from 40 to 70\,N ($p\leq0.009$ at each level). Matched simulation showed nearly identical steady-contact working-shoulder RMS speeds for SHS and Generic (0.0319 and 0.0320\,m\,s$^{-1}$; $p=0.919$; Fig.~\ref{fig:simulation}(e)). The improved hardware repeatability was therefore not explained by SHS simply moving the working shoulder more slowly.

\subsection{Transfer Across Task Regions Without Retraining}

\begin{figure}[t]
  \centering
  \includegraphics[width=\linewidth]{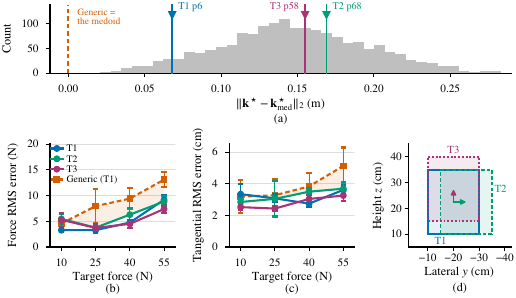}
  \caption{SHS transfer without retraining from T1 to task regions translated by 5\,cm laterally (T2) and vertically (T3). (a) WBKR distances from the training-set medoid over 1,680 training configurations, with the optimised T1--T3 references marked by percentile. (b,c) Normal-force and tangential-position RMS errors (mean $\pm$ SD); lower is better. (d) Task-region geometry.}
  \label{fig:transfer}
\end{figure}

Three different task-region geometries are shown in Fig.~\ref{fig:transfer}(d). Individual re-optimisation produced different support configurations and hence different WBKR references. The T1 reference was close to the Generic reference ($d_{\mathrm{med}}=0.068$\,m); T2 and T3 were farther away (0.169 and 0.155\,m). Measured by $d_{\mathrm{med}}$, T1, T2, and T3 are farther from the Generic reference than 6\%, 68\%, and 58\% of the training-set references, respectively (Fig.~\ref{fig:transfer}(a)).

Hardware tracking errors remained within a narrow range across the three task regions: over the 10--55\,N range, force RMS errors were 5.2--6.2\,N and tangential-position RMS errors were 2.8--3.4\,cm, as shown in Fig.~\ref{fig:transfer}(b,c). Thus, the same policies executed the shifted tasks using newly optimised support configurations without retraining.

\section{Discussion}

Establishing environmental bracing substantially expands the achievable force range. SHS sustains contact and tracks target normal forces over a wide range that neither of the tested controllers without environmental bracing can cover reliably. The two methods without environmental bracing fail at opposite ends of the force range. Modular keeps an upright stance and cannot maintain the forward lean required to balance large reaction forces, limiting its maximum usable force. In contrast, No-bracing adopts a forward lean, but without a second contact the lean load must pass through the working hand, imposing a non-zero minimum force. Environmental bracing overcomes both limitations and therefore widens the usable force range.

Within this widened force range, task conditioning further improved tracking, and the benefit was robust across the 30 matched simulation seeds. Compared with Generic, SHS improved force accuracy, episode success rate, and contact retention, while contact-window position MAE remained similar. On real hardware, SHS also reduced normal-force RMS error and, under higher loads, tangential-position RMS error, showing that the task-conditioned support configuration improves both force and position tracking during forceful manipulation. Generic additionally showed larger working-shoulder wobble at high force, while shoulder speed remained similar in matched simulation. This suggests that improved working-shoulder repeatability may contribute to the tracking improvement.

Despite these improvements, SHS requires the entire task region to be reachable by the working arm from a single optimised support configuration. Evaluation was limited to one planar support surface. Different contact planes, surface orientations, curved surfaces, and supporting-hand grasps or hooks remain untested. Such contacts could provide tensile as well as compressive and frictional support, enabling a wider range of support configurations.

\section{Conclusion}

We introduced SHS, which uses task-conditioned environmental bracing to extend forceful manipulation for floating-base humanoids without human motion data or online whole-body trajectory planning. On a Unitree G1, SHS achieved usable forces up to 60\,N, whereas the unsupported baseline never sustained more than 13.5\,N for one second. Task-conditioned bracing improved tracking relative to a task-independent reference, and the same frozen policies executed shifted task regions without retraining. These results demonstrate that deliberately bracing against the environment can substantially extend the forceful manipulation capability of humanoid robots.

\ifRALanonymous\else
\section*{Acknowledgment}
This work was supported by the Australian Research Council under Grant No. IC200100001. The authors acknowledge the support of Queensland University of Technology, the Australian Cobotics Centre, and the QUT Centre for Robotics.
\fi

\bibliographystyle{IEEEtran}
\bibliography{reference}

\end{document}